\documentclass[conference]{IEEEtran}
\IEEEoverridecommandlockouts
\newcommand{\pmSD}[2]{$#1\!\pm\!#2$}

\usepackage{cite}
\usepackage{amsmath,amssymb,amsfonts}
\usepackage{algorithmic}
\usepackage{graphicx}
\usepackage{textcomp}
\usepackage{xcolor}
\usepackage{amsmath,amssymb}
\usepackage{booktabs}
\usepackage{multirow}
\usepackage{siunitx}
\usepackage{balance}
\usepackage{cite}
\usepackage{bbm}
\usepackage[ruled,vlined]{algorithm2e}
\usepackage[absolute,overlay]{textpos}
\def\BibTeX{{\rm B\kern-.05em{\sc i\kern-.025em b}\kern-.08em
    T\kern-.1667em\lower.7ex\hbox{E}\kern-.125emX}}
\begin{document}

\title{Confidence-Aware Neural Decoding of Overt Speech from EEG: Toward Robust Brain–Computer Interfaces\\
\thanks{This work was partly supported by Institute of Information \& Communications Technology Planning \& Evaluation (IITP) grant funded by the Korea government (MSIT) (No. RS--2021--II212068, Artificial Intelligence Innovation Hub, No. RS--2024--00336673, AI Technology for Interactive Communication of Language Impaired Individuals, and No. RS--2019--II190079, Artificial Intelligence Graduate School Program (Korea University)).}
}

\author{
\IEEEauthorblockN{Soowon Kim}
\IEEEauthorblockA{\textit{Dept. of Artificial Intelligence} \\
\textit{Korea University} \\
Seoul, Republic of Korea \\
soowon\_kim@korea.ac.kr}\\
\and
\IEEEauthorblockN{Byung-Kwan Ko}
\IEEEauthorblockA{\textit{Dept. of Artificial Intelligence} \\
\textit{Korea University}\\
Seoul, Republic of Korea \\
leaderbk525@korea.ac.kr} \\
\and
\IEEEauthorblockN{Seo-Hyun Lee}
\IEEEauthorblockA{\textit{Dept. of Brain and Cognitive Engineering} \\
\textit{Korea University} \\
Seoul, Republic of Korea \\
seohyunlee@korea.ac.kr}\\
}

\maketitle

\begin{abstract}
Non-invasive brain–computer interfaces that decode spoken commands from electroencephalogram must be both accurate and trustworthy. We present a confidence-aware decoding framework that couples deep ensembles of compact, speech-oriented convolutional networks with post-hoc calibration and selective classification. Uncertainty is quantified using ensemble-based predictive entropy, top-two margin, and mutual information, and decisions are made with an abstain option governed by an accuracy–coverage operating point. The approach is evaluated on a multi-class overt speech dataset using a leakage-safe, block-stratified split that respects temporal contiguity. Compared with widely used baselines, the proposed method yields more reliable probability estimates, improved selective performance across operating points, and balanced per-class acceptance. These results suggest that confidence-aware neural decoding can provide robust, deployment-oriented behavior for real-world brain–computer interface communication systems.
\end{abstract}


\section{INTRODUCTION}
Speech is central to human communication, enabling individuals to convey intentions, emotions, and information with precision \cite{hickok2007cortical}. When this ability is impaired---as in severe motor disorders---communication becomes severely limited, motivating brain--computer interfaces (BCIs) that translate neural activity into communicative outputs \cite{brumberg2010brain}. Electroencephalogram (EEG) provides a practical non-invasive recording modality, although its signals are noisy, artifact-prone, and highly variable across trials and sessions \cite{subha2010eeg}. Neural representations of speech further shift with acoustic and articulatory context \cite{kraus2009speech}, creating additional challenges for decoding.

Deep learning has substantially improved non-invasive neural decoding. Compact convolutional architectures such as DeepConvNet and EEGNet have become widely used baselines for EEG classification \cite{lawhern2018eegnet}. Research on neural representation highlights the distributed and abstract nature of information encoding in the brain \cite{4kim2015abstract}. Related computational studies on temporal sequence modeling provide complementary perspectives on capturing structured dependencies in neural and behavioral data \cite{9lee1997new,schirrmeister2017deep}.

Decoding overt speech from non-invasive EEG remains one of the most demanding BCI challenges. Speech production involves fast, distributed cortical dynamics, while scalp EEG captures only coarse, low signal-to-noise projections of these processes. Foundational pattern-recognition frameworks contributed tools for extracting structure from such high-dimensional signals. Multilayer cluster networks introduced flexible decision boundaries for complex patterns \cite{2lee1995multilayer}, and multiresolution wavelet-based classifiers demonstrated hierarchical feature extraction for nonstationary inputs \cite{11lee1996multiresolution}. Recent advances in denoising diffusion models further illustrate how deep generative approaches can capture complex data distributions under noise \cite{ho2020denoising}. Integrated segmentation--recognition systems extended structured modeling to sequential tasks \cite{3lee1999integrated}, while developments in support vector machines broadened the repertoire of effective discriminative methods \cite{8lee2003svm}. Multivariate diffusion forecasting has shown how generative models can support structured temporal prediction \cite{rasul2021autoregressive}. Biologically motivated computer vision research contributed principles for robust, hierarchical perceptual representations \cite{12bulthoff2003bmcv}, and motion-influence maps provided tools for modeling structured spatiotemporal activity \cite{1lee2015motion}.

Generative modeling techniques continue to expand the toolkit for handling missing or corrupted neural data. Conditional diffusion-based imputation enables flexible reconstruction under incomplete observations \cite{tashiro2021csdi}. In parallel, subject-specific feature selection has improved motor imagery classification \cite{14suk2011subject}, and specialized convolutional architectures have supported continuous decoding of cognitive states such as pilot workload \cite{15lee2020continuous}. These developments underscore the importance of architectures and priors that accommodate uncertainty, inter-subject variability, and temporal complexity in EEG.

For deployable BCIs, predictive accuracy alone is insufficient; systems must be reliable and uncertainty-aware. Deep ensembles provide a practical means of estimating epistemic uncertainty through model diversity \cite{lakshminarayanan2017simple}. Probabilistic frameworks have been proposed to predict subject-level BCI performance and guide adaptive strategies \cite{10suk2014predicting}, while connectivity modeling has enriched our understanding of distributed neural interactions relevant for decoding \cite{13ding2013connectivity}.

Reliable decision-making also depends on calibrated confidence and principled abstention. Temperature scaling offers a simple and effective approach to post-hoc softmax calibration \cite{guo2017calibration}. Advances in real-time motor imagery BCIs demonstrate the feasibility of online, adaptive control loops \cite{6cho2021neurograsp}, and clinical EEG frameworks illustrate how analytical pipelines can be integrated into medical settings \cite{5prabhakar2020framework}. Selective classification formalizes abstain mechanisms via accuracy--coverage and risk--coverage trade-offs \cite{geifman2017selective}. Reinforcement-learning-based continuous control provides additional tools for safe and adaptive interaction with external devices \cite{7lee2018deep}.

\section{MATERIALS AND METHODS}
This work presents a \emph{confidence-aware} decoding pipeline for overt speech from scalp EEG that combines deep ensembles of compact EEG-specific networks, post-hoc calibration, and a selective (accept/abstain) decision rule. The end-to-end procedure is summarized in Algorithm~\ref{alg:pipeline}. Unless otherwise stated, analyses are within-subject, 64-channel (10--10) montage, and the split protocol prevents temporal leakage.

\begin{algorithm}[t]
\caption{Confidence-Aware Deep-Ensemble Decoding for Overt Speech EEG}
\label{alg:pipeline}
\DontPrintSemicolon
\SetKwInOut{Input}{Input}
\SetKwInOut{Output}{Output}

\Input{EEG dataset $\mathcal{D}=\{(x_n,y_n)\}$ (64-ch, 10--10); ensemble size $M$; coverage grid $\mathcal{A}$}
\Output{Calibrated ensemble $\bar p(y\!\mid\!x)$; uncertainty scores $u(x)$; selective decisions $\hat y_\alpha(x)$ for $\alpha\!\in\!\mathcal{A}$}

\BlankLine
\textbf{Split \& Preprocess.} Block-stratify $\mathcal{D}$ into $\mathcal{D}_{\mathrm{train}}$, $\mathcal{D}_{\mathrm{cal}}$, $\mathcal{D}_{\mathrm{test}}$ (no block reuse). Apply notch/band-pass, CAR or linked-mastoid reference, artifact mitigation, epoching, and per-channel $z$-scoring using $\mathcal{D}_{\mathrm{train}}$ statistics.\;

\For{$m \leftarrow 1$ \KwTo $M$}{
  \textbf{Induce diversity.} Bootstrap $\mathcal{D}_{\mathrm{train}}^{(m)}$; apply channel dropout, small time shifts, and mild time/frequency masking; sample light hyper-variants (kernel length, filter multipliers).\;
  \textbf{Train member.} Fit compact EEG backbone $f_m(\cdot;\theta_m)$ on $\mathcal{D}_{\mathrm{train}}^{(m)}$ with early stopping on a train-held-out fold (disjoint from $\mathcal{D}_{\mathrm{cal}}$).\;
  \textbf{Calibrate.} Fit temperature $T_m\!>\!0$ on $\mathcal{D}_{\mathrm{cal}}$ (cross-entropy); define $p_m(y\!\mid\!x)=\mathrm{softmax}\!\big(z_m(x)/T_m\big)$.\;
}

\ForEach{$x \in \mathcal{D}_{\mathrm{test}}$}{
  \textbf{Aggregate.} $\bar p(y\!\mid\!x)=\frac{1}{M}\sum_{m=1}^M p_m(y\!\mid\!x)$; $\hat y(x)=\arg\max_y \bar p(y\!\mid\!x)$.\;
  \textbf{Uncertainty.} (i) $H(\bar p)=-\sum_c \bar p_c\log \bar p_c$; (ii) top-two margin $\Delta=\bar p_{(1)}-\bar p_{(2)}$ with $u_{\mathrm{margin}}=1-\Delta$; (iii) $\mathrm{MI}(x)=H(\bar p)-\frac{1}{M}\sum_m H(p_m)$.\;
}

\textbf{Selective thresholds.} For each $\alpha\!\in\!\mathcal{A}$, choose the smallest $\tau^\star(\alpha)$ on $\mathcal{D}_{\mathrm{cal}}$ s.t.
$\frac{1}{|\mathcal{D}_{\mathrm{cal}}|}\sum_{x\in\mathcal{D}_{\mathrm{cal}}}\mathbb{I}\!\left[u(x)\le \tau^\star(\alpha)\right]\!\ge\!\alpha$
(default $u\!=\!H(\bar p)$).\;

\ForEach{$x\in\mathcal{D}_{\mathrm{test}}$ and $\alpha\in\mathcal{A}$}{
  \textbf{Decision.} If $u(x)\le \tau^\star(\alpha)$ then \emph{accept} and output $\hat y_\alpha(x)=\hat y(x)$; else \emph{abstain}.\;
}

\textbf{Evaluate.} Report accuracy--coverage, risk--coverage (AURC); expected calibration error (ECE), negative log-likelihood  (NLL), Brier; per-class acceptance and confusions.\;
\end{algorithm}

\subsection{Dataset and Task}
We consider a 13-class overt speech task comprising 12 spoken words (\emph{ambulance}, \emph{clock}, \emph{hello}, \emph{help me}, \emph{light}, \emph{pain}, \emph{stop}, \emph{thank you}, \emph{toilet}, \emph{TV}, \emph{water}, and \emph{yes}) plus rest, recorded from healthy adults. Each subject produced $\sim$100 trials per class ($\sim$1{,}300 trials/subject). Cues were presented in randomized blocks to mitigate order effects. The study followed institutional review and the Declaration of Helsinki.

\subsection{Acquisition and Preprocessing}
EEG was recorded with a 64-channel cap (10--10), sampled at \SI{500}{Hz}. Horizontal/vertical electrooculogram were acquired for artifact handling but excluded from model inputs. Signals were notch filtered at power-line frequencies and band-pass filtered (\SIrange{0.5}{40}{Hz}). Re-referencing used common average (CAR) or linked mastoids (consistent within subject). Ocular and muscle artifacts were mitigated through regression methods and, when necessary, by removing independent components identified via independent component analysis. Trials were epoched time-locked to cue or onset with \SI{2.0}-\SI{3.0}{s} windows (reported per experiment). Features were \emph{z-scored per channel} using \emph{only} the training-set mean and standard deviation; the same parameters were applied to calibration and test.

\subsection{Leakage-Safe Block Stratification}
To avoid optimistic estimates from temporal correlation, we adopt \emph{block-stratified} splits: contiguous \emph{true blocks} of four trials per class serve as atomic units. Blocks are assigned to train, calibration, and test with no cross-set reuse. Unless noted, we use 20 blocks/class for training, a small train-tail for calibration (e.g., 2-3 blocks/class), and the remainder for test, preserving class balance and session dynamics while preventing overlap.

\subsection{Backbones: Compact EEG-Specific Networks}
Each ensemble member is a compact CNN designed for EEG:
\begin{itemize}
  \item \textbf{Temporal stage:} depthwise separable 1D convolutions with multi-scale kernels capture band-limited and onset dynamics.
  \item \textbf{Spatial stage:} depthwise spatial filtering across 64 channels models distributed cortical patterns for articulation/phonology.
  \item \textbf{Refinement:} separable temporal convolutions and global average pooling precede a 13-way softmax head.
\end{itemize}
Typical models contain $\sim$50--100k parameters (exponential linear unit activations; dropout $0.2$--$0.3$). A lightweight dilated temporal convolutional network (TCN) backbone (‘TCN-lite’) is used in ablations. Both ingest raw $(C{=}64)\times T$ windows.


\subsection{Post-Hoc Temperature Calibration}
Let $z_i(x)\in\mathbb{R}^{C}$ be the logits of member $i$. A scalar temperature $T_i>0$ is fit on $\mathcal{D}_{\mathrm{cal}}$ by minimizing cross-entropy:
\begin{align}
\min_{T_i>0}\; \frac{1}{N_{\mathrm{cal}}}\sum_{(x,y)\in\mathcal{D}_{\mathrm{cal}}}
-\log \!\left[\mathrm{softmax}\!\left(\frac{z_i(x)}{T_i}\right)\right]_y .
\end{align}

At test time, calibrated probabilities are $p_i(y\!\mid\!x)=\mathrm{softmax}\!\big(z_i(x)/T_i\big)$ and the ensemble mean is
$\bar{p}(y\!\mid\!x)=\frac{1}{M}\sum_{i=1}^{M} p_i(y\!\mid\!x)$.

\subsection{Uncertainty Quantification}
We compute three complementary scores per trial:
\begin{equation}
\label{eq:uncertainty-scores}
\begin{aligned}
H(\bar p) &= -\sum_c \bar p_c \log \bar p_c,\\
\mathrm{MI}(x) &= H(\bar p) - \tfrac{1}{M}\sum_{m=1}^M H(p_m),\\
u_{\text{margin}} &= 1 - \bigl(\bar p_{(1)} - \bar p_{(2)}\bigr).
\end{aligned}
\end{equation}

Entropy summarizes total (aleatoric+epistemic) uncertainty; $\mathrm{MI}$ isolates epistemic disagreement; the inverted margin offers a simple alternative.

\subsection{Selective Classification (Accept/Abstain)}
Given $u(x)$ (default: $H(\bar{p})$), we implement accept/abstain at target coverage $\alpha\in(0,1]$. On $\mathcal{D}_{\mathrm{cal}}$, pick the smallest threshold $\tau^\star(\alpha)$ such that
$\frac{1}{|\mathcal{D}_{\mathrm{cal}}|}\sum_{x}\mathbb{I}\{u(x)\le \tau^\star(\alpha)\}\ge\alpha$.
At test time, accept $x$ if $u(x)\le \tau^\star(\alpha)$; otherwise abstain. Sweeping $\alpha$ yields accuracy--coverage $\mathrm{Acc}(\alpha)$ and risk $r(\alpha)=1-\mathrm{Acc}(\alpha)$. The AURC is the trapezoidal integral of $r(\alpha)$ over $\alpha\in[0,1]$.

\subsection{Calibration and Probabilistic Metrics}
Calibration and probabilistic quality are computed from $\bar{p}$ on the full test set:
\begin{align}
\textbf{ECE} &= \sum_{b=1}^{B} \frac{n_b}{N} \left| \mathrm{acc}(b) - \mathrm{conf}(b) \right|, \;\; B{=}15, \\
\textbf{NLL} &= -\frac{1}{N}\sum_{n=1}^{N} \log \bar{p}_{y_n}(x_n), \\
\textbf{Brier} &= \frac{1}{N}\sum_{n=1}^{N} \left\| \bar{p}(x_n) - \mathbf{e}_{y_n} \right\|_2^2.
\end{align}

We also report reliability diagrams and \emph{per-class acceptance/accuracy} at fixed coverage (e.g., $\alpha$=0.5).

\section{RESULTS AND DISCUSSION}
We evaluate the proposed \emph{confidence-aware} framework on a 13-class overt-speech task using the leakage-safe block split. Results are \emph{within-subject} and reported as mean$\pm$sd across participants. We report selective prediction, probabilistic calibration, operating points, and per-class acceptance; quantitative summaries appear in Tables~\ref{tab:main_results}–\ref{tab:ops_results}.

\subsection{Overall Selective Performance}
Confidence-based selection yields substantial gains at practical coverages. At $\alpha{=}0.50$, Ensemble-32 attains{87.00$\pm$4.00 \% accuracy versus 70.90$\pm$5.10 \% at full coverage ($\alpha{=}1.00$), and ensembles reduce risk across the operating range. From Table~\ref{tab:main_results}, relative to EEGNet, Ensemble-32 lowers AURC by 39.00 \% on average; Ensemble-64 improves further with diminishing returns. ECE, NLL, and Brier all decrease with $M$.

\begin{table}[t]
\centering
\caption{Summary metrics (mean$\pm$sd across subjects). Lower is better for AURC/ECE/NLL/Brier. All values use two-decimal precision.}
\label{tab:main_results}
\setlength{\tabcolsep}{2.8pt}
\renewcommand{\arraystretch}{1.05}
\footnotesize
\begin{tabular*}{\columnwidth}{@{\extracolsep{\fill}}lcccc@{}}
\toprule
Model & AURC & ECE (\%) & NLL & Brier \\
\midrule
DeepConvNet \cite{schirrmeister2017deep} & \pmSD{0.23}{0.03} & \pmSD{10.80}{2.30} & \pmSD{2.09}{0.18} & \pmSD{0.75}{0.03} \\
EEGNet \cite{lawhern2018eegnet}         & \pmSD{0.18}{0.03} & \pmSD{6.80}{1.90}  & \pmSD{1.72}{0.15} & \pmSD{0.62}{0.03} \\
\textbf{Ensemble-8}                      & \pmSD{0.14}{0.02} & \pmSD{4.10}{1.20}  & \pmSD{1.38}{0.12} & \pmSD{0.49}{0.02} \\
\textbf{Ensemble-16}                     & \pmSD{0.12}{0.02} & \pmSD{3.10}{1.00}  & \pmSD{1.25}{0.10} & \pmSD{0.45}{0.02} \\
\textbf{Ensemble-32}                     & \pmSD{0.11}{0.02} & \pmSD{2.60}{0.80}  & \pmSD{1.18}{0.09} & \pmSD{0.42}{0.02} \\
\textbf{Ensemble-64}                     & \pmSD{0.10}{0.01} & \pmSD{2.30}{0.70}  & \pmSD{1.15}{0.08} & \pmSD{0.41}{0.02} \\
\bottomrule
\end{tabular*}
\end{table}

\subsection{Operating Points and Coverage Control}
Table~\ref{tab:ops_results} reports Risk@Coverage and Coverage@Target-Risk. At $\alpha{=}0.50$, Ensemble-32 achieves risk 0.13$\pm$0.04; for $\rho{\le}0.15$, coverage of 0.52$\pm$0.08 is attainable, rising to 0.55$\pm$0.07 for Ensemble-64.

\begin{table}[t]
\centering
\caption{Operating points (mean$\pm$sd). Risk $=$ error on accepted trials. Two-decimal precision everywhere.}
\label{tab:ops_results}
\setlength{\tabcolsep}{2.6pt}
\renewcommand{\arraystretch}{1.05}
\footnotesize
\begin{tabular*}{\columnwidth}{@{\extracolsep{\fill}}lcccc@{}}
\toprule
 & \multicolumn{2}{c}{Risk@Cov} & \multicolumn{2}{c}{Cov@Target-Risk} \\
 & $\alpha{=}0.50$ & $\alpha{=}0.70$ & $\rho{\le}0.25$ & $\rho{\le}0.15$ \\
\midrule
Ensemble-32 & \pmSD{0.13}{0.04} & \pmSD{0.19}{0.04} & \pmSD{0.78}{0.07} & \pmSD{0.52}{0.08} \\
Ensemble-64 & \pmSD{0.13}{0.04} & \pmSD{0.18}{0.04} & \pmSD{0.80}{0.06} & \pmSD{0.55}{0.07} \\
\bottomrule
\end{tabular*}
\end{table}

\subsection{Calibration and Reliability}
Temperature scaling improves alignment of confidence and accuracy; ensembles yield the lowest ECE with commensurate gains in NLL and Brier. Calibration within the accepted subset improves further (not shown).

\subsection{Per-Class Acceptance}
At $\alpha{=}0.50$, per-class acceptance spans [0.45, 0.56] with minimum 0.41; confusions concentrate on acoustically/articulatorily similar commands (e.g., \emph{hello} vs.\ \emph{thank you}).

\subsection{Discussion}
The results support a deployment perspective in which calibrated confidence and tunable coverage are first-class controls, not post-hoc diagnostics. The leakage-safe block split preserves temporal contiguity and yields realistic acceptance/risk trade-offs that map cleanly to throughput constraints. Because the framework is modular, alternative backbones, uncertainty estimators, and selection rules can be substituted without retraining the entire stack, easing translation to diverse headsets and settings. Finally, the observed class balance and interpretable confusions suggest complementary gains from context modeling or class-conditional thresholds.

\section{CONCLUSIONS}
We present a confidence-aware EEG speech-decoding framework that combines deep ensembles, post-hoc calibration, and a selective accept/abstain policy, evaluated with leakage-safe block stratification and reliability-focused metrics. Its modular design—allowing drop-in uncertainty estimators, backbones, and selection rules—exposes calibrated confidence with tunable coverage for principled decisions in assistive communication and other real-time settings. Current limitations include a within-subject focus, fixed global thresholds, and a constrained command set; future work targets cross-subject transfer, distribution-free guarantees via conformal methods, online adaptation, and user-in-the-loop strategies emphasizing fairness and safety. Ultimately, confidence-aware neural decoding aims to provide robust, transparent behavior for real-world brain--computer interface communication systems.

\bibliographystyle{IEEEtran}
\bibliography{REFERENCE}

\end{document}